\documentclass[11pt]{article}
\usepackage{graphicx} 
\usepackage{amsmath}  
\usepackage{siunitx}  
\usepackage{authblk}
\usepackage{caption}
\usepackage{makecell}
\usepackage{subcaption}
\usepackage{array}
\usepackage{multirow}
\usepackage[margin=1in]{geometry}
\usepackage[backend=biber,style=apa]{biblatex}
\title{Axis-Aligned 3D Stalk Diameter Estimation from RGB-D Imagery}
\author{Benjamin Vail}
\author{Rahul Harsha Cheppally}
\author{Ajay Sharda}
\author{Sidharth Rai}
\affil{Department of Biological and Agricultural Engineering, Kansas State University, Manhattan, KS 66506; USA}
\date{July 2025}

\begin{document}

\maketitle

\section{Introduction}

With the global population projected to reach 9.7 billion by 2050 (\cite{unitednationsdepartmentofeconomicandsocialaffairsundesaWorldTotalPopulation2024}), the challenge of ensuring food security will require major advances in agricultural productivity and resilience. Precision phenotyping—a data-driven approach to crop breeding and management—offers a powerful tool for achieving this goal by enabling high-resolution, trait-specific insights across large populations.

Among structural traits, stalk diameter plays a critical role in crop performance, influencing factors such as lodging resistance, biomass allocation, hydraulic function, and disease vulnerability (\cite{robertsonMaizeLodgingResistance2022, wakamoriOpticalFlowBasedAnalysis2019, xueEvaluationMaizeLodging2020}). Accordingly, it is an important metric for both breeding programs and agronomic monitoring. However, conventional measurement techniques using calipers or static rigs are labor-intensive, inconsistent, and poorly suited to the demands of modern high-throughput phenotyping or in-field crop monitoring.

Recent advances in deep learning-based computer vision have demonstrated impressive capabilities in complex agricultural environments—such as detecting seeds in noisy conditions (\cite{sls}) and detecting green apple in highly complex environments (\cite{rfdetr}). These developments suggest strong potential for applying similar methods to the challenge of stalk segmentation and measurement under real-world, occluded under-canopy conditions.

In this study, we present a geometry-aware computer vision pipeline for estimating stalk diameter from RGB-D images. Our method combines deep learning-based instance segmentation, 3D point cloud reconstruction, and axis-aligned slicing using Principal Component Analysis (PCA) to determine the stalk’s dominant orientation. By performing cross-sectional measurements orthogonal to this axis, we improve the accuracy and robustness of diameter estimation—even in the presence of stalk curvature, arbitrary orientation, and field-induced noise. This approach is designed for scalable deployment, offering a viable pathway toward automated, high-throughput phenotyping and adaptive crop management.

\section{Background}

Stalk diameter is a well-studied trait in crop phenotyping due to its correlation with mechanical stability, biomass production, and lodging resistance (\cite{robertsonMaizeLodgingResistance2022, xueEvaluationMaizeLodging2020}). It has also been associated with physiological traits such as hydraulic conductivity and vulnerability to disease-related wilting (\cite{wakamoriOpticalFlowBasedAnalysis2019}). Consequently, the accurate and repeatable measurement of stalk diameter is of considerable interest in both research and applied agricultural settings.

Historically, diameter measurements have been obtained using manual tools like calipers or static imaging rigs, which are labor-intensive and impractical for high-throughput applications. With the increasing availability of RGB-D sensors, there has been growing interest in automating stalk measurement using 3D data. However, field deployment of such systems faces numerous obstacles: depth sensors are sensitive to ambient light and wind, and may produce incomplete or noisy measurements of slender targets like plant stems.

Recent work has attempted to extract stalk diameters from RGB-D images using methods such as color thresholding, edge detection, or rule-based segmentation (\cite{erndweinFieldbasedMechanicalPhenotyping2020}). These methods often assume vertical stalk alignment or use simplified geometric fitting, which limits their reliability in complex or variable environments. More advanced approaches like PST (\cite{duPSTPlantSegmentation2023}) and other instance segmentation-based pipelines (\cite{charisisDeepLearningbasedInstance2024}) have improved stalk detection and 3D reconstruction, but they frequently lack robust axis alignment or are not designed to generalize across different species or experimental setups.

To overcome these limitations, our pipeline incorporates deep learning-based instance segmentation to isolate individual stalks in RGB images, followed by 3D reconstruction using camera intrinsics and depth data. We use Principal Component Analysis (PCA) calculated using Singular Value Decomposition (SVD) to determine each stalk's dominant axis, enabling diameter estimation via cross-sectional slicing orthogonal to this axis. This strategy improves measurement accuracy and enhances robustness under diverse imaging conditions.

\section{Methodology}

\subsection*{Overview}
This study presents a computer vision and 3D reconstruction system developed to measure plant stalk diameter from RGB-D image pairs. The system combines instance segmentation, depth-based point cloud reconstruction, spatial filtering, and geometric modeling to estimate stalk diameters with high precision. Each stage of the pipeline is detailed below (see Fig. 1).

\begin{figure}[htbp]
    \centering
    \includegraphics[width=0.95\textwidth]{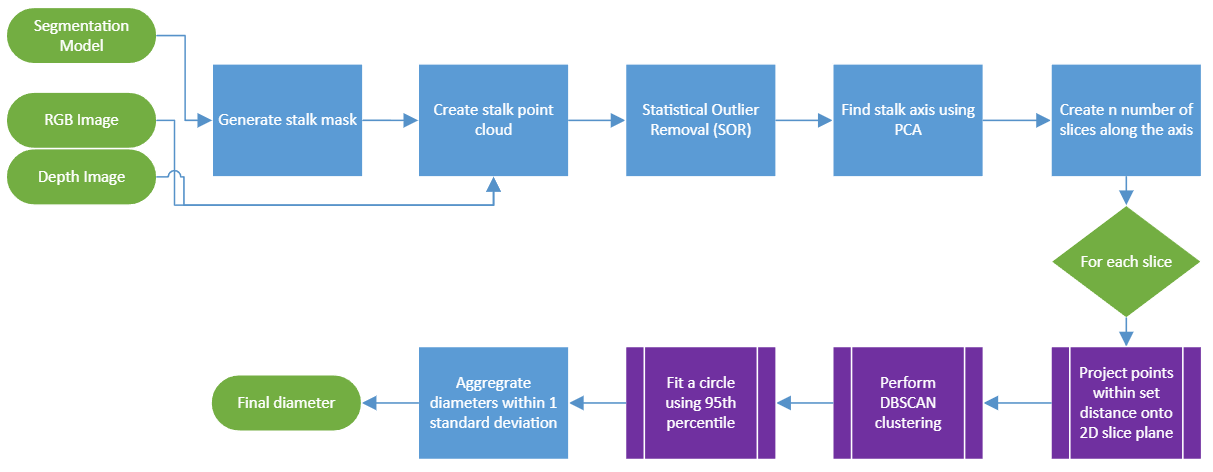}
    \caption{Flowchart of the pipeline}
    \label{fig:pipeline_flowchart}
\end{figure}

\subsection*{Data Acquisition and Calibration}
RGB and depth images were captured using a Luxonis OAK-D short-range (SR) stereo camera (\cite{luxonis_oak-d_2025}), which was chosen for its optimization for near-field depth perception. The SR model features a reduced baseline—the distance between the two stereo cameras—which lowers the minimum distance for accurate depth perception. This characteristic is critical for measuring small, nearby objects like plant stalks, as it ensures reliable depth data at close proximity. Using a pinhole camera model, the camera's intrinsic parameters, including focal lengths and principal point coordinates, were used to project image data into 3D space. Depth measurements were then scaled to metric units to create the final 3D point cloud.

The training dataset was created using artificial plants with clearly visible stalk sections to facilitate algorithm testing. Images were captured under varied lighting conditions and at varying distances. After applying flip, rotation, exposure, and noise augmentations, the dataset totaled 93 images with an 80/15/5 train/validation/test split. The small dataset size is due to the uniformity of the artificial plants used for lab testing. Future work will require a much larger and more varied real-plant dataset for in-situ deployment.

\subsection*{Instance Segmentation}
An instance segmentation model was trained to isolate individual plant stalks from the background. Segmentation was performed using a YOLOv11x-seg model (\cite{jocher_ultralytics_2024}), which was pre-trained on the Common Objects in Context (COCO) dataset (\cite{lin_microsoft_2015}). The model was subsequently fine-tuned for this application via transfer learning on a custom dataset of artificial plant stalks.

For the fine-tuning process, the Adam optimizer was selected for its effective performance with smaller training datasets. Training was scheduled for 300 epochs with an early stopping mechanism to prevent overfitting; training concluded at epoch 186 when no improvement in validation loss was observed. The resulting model achieved a mean Average Precision (mAP50) of 0.817, a precision of 0.910, and a recall of 0.739 on the validation set (see Fig. 2).

\begin{figure}[htbp]
    \centering
    \includegraphics[width=0.6\linewidth]{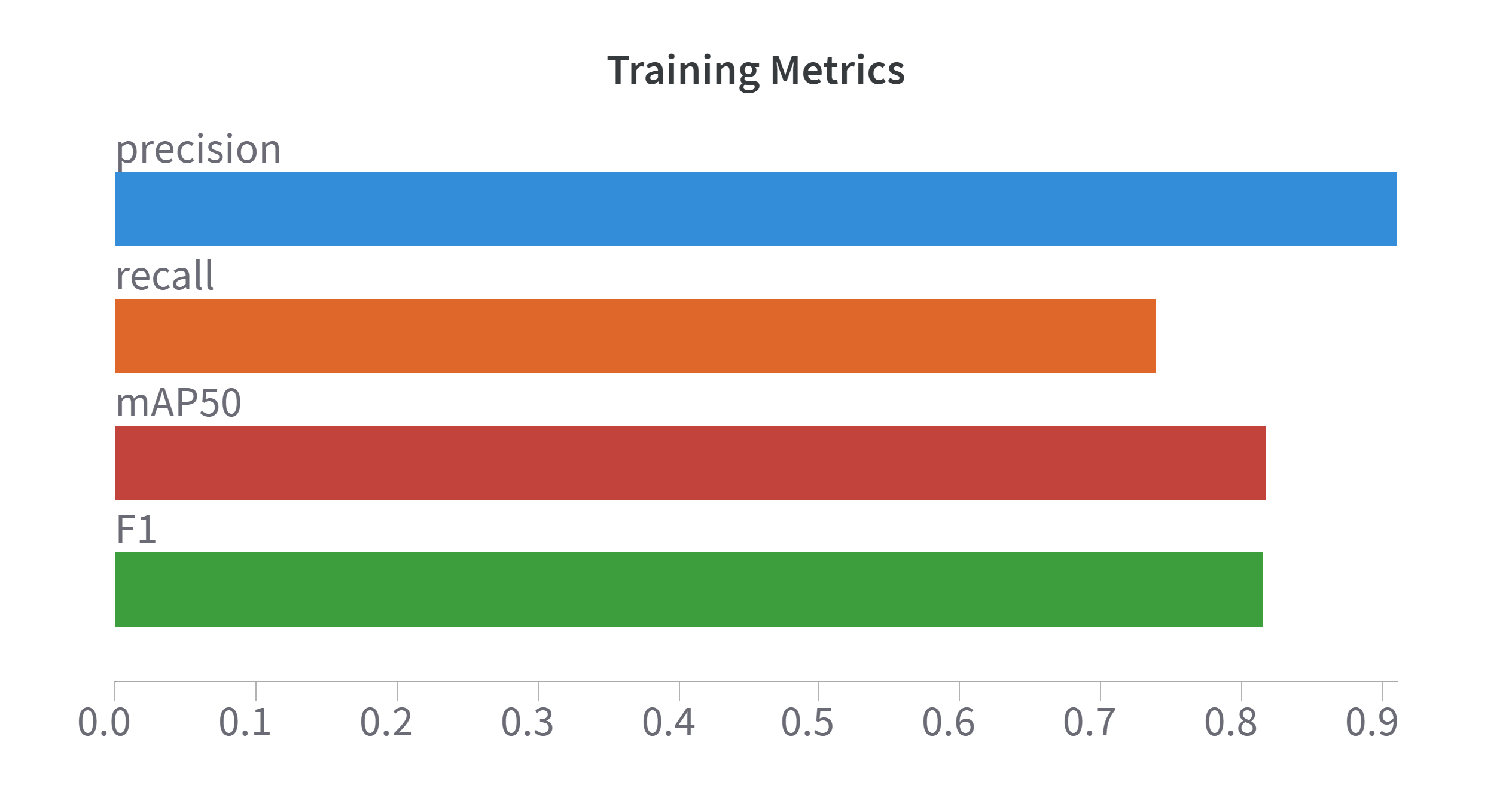}
    \caption{Model Statistics}
    \label{fig:model_stats}
\end{figure}

When applied to an RGB image, the model produces a pixel-wise segmentation mask for each detected object. For each valid detection, the corresponding mask was resized to match the depth image resolution and then binarized. This process yields a final binary foreground mask that spatially defines a single stalk instance for subsequent 3D analysis (see Fig. 3). Although this model's performance was sufficient for the controlled environment of this experiment, future in-field implementation will require a more robust model.

\begin{figure}[htbp]
    \centering
    \begin{subfigure}[b]{0.48\textwidth}
        \centering
        \includegraphics[width=\linewidth]{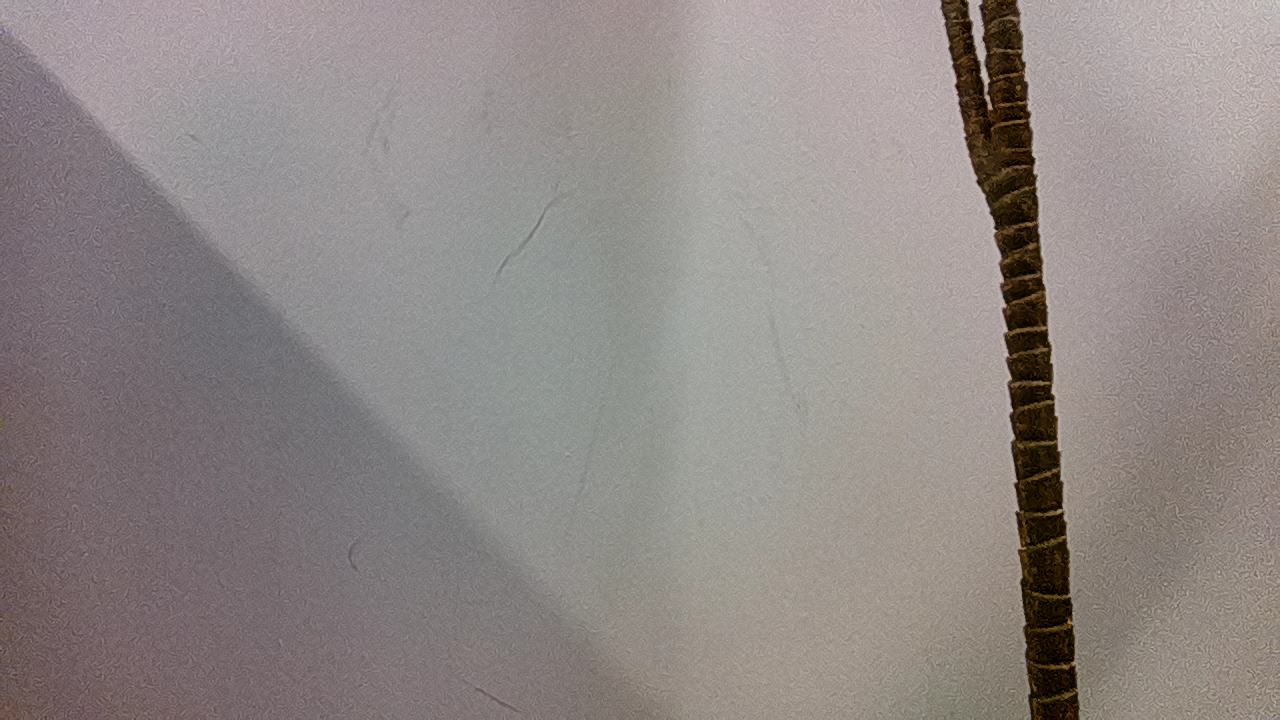}
        \caption{Artificial plant stalk used for validation.}
        \label{fig:artificial_plant}
    \end{subfigure}
    \hfill
    \begin{subfigure}[b]{0.48\textwidth}
        \centering
        \includegraphics[width=\linewidth]{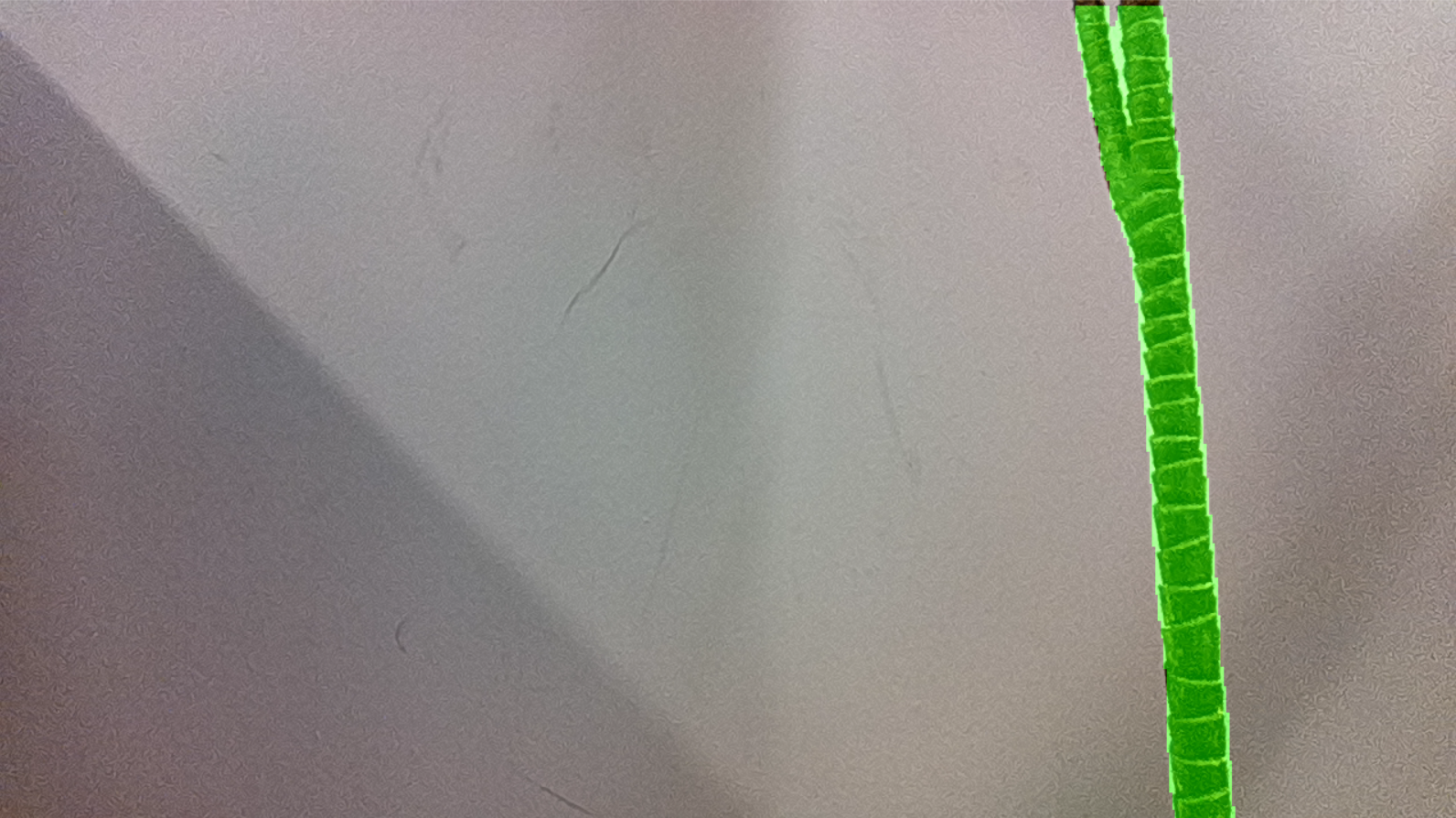}
        \caption{The stalk instance segmentation mask.}
        \label{fig:segmentation_mask}
    \end{subfigure}
    \caption{Instance segmentation applied to an artificial plant stalk.}
    \label{fig:segmentation_example}
\end{figure}

\subsection*{3D Point Cloud Generation}
A masked RGB-D image was created by applying the binary mask to both the color and depth channels. The Open3D library (\cite{zhou_open3d_2018}) was then used to construct a colored 3D point cloud from the masked image using the camera's intrinsic parameters. The resulting point cloud was filtered using a Statistical Outlier Removal (SOR) technique with six nearest neighbors and a standard deviation threshold of 4.0 to improve geometric consistency and reduce noise (see Fig. 3).

\begin{figure}[htbp]
    \centering
    \includegraphics[width=0.2\textwidth]{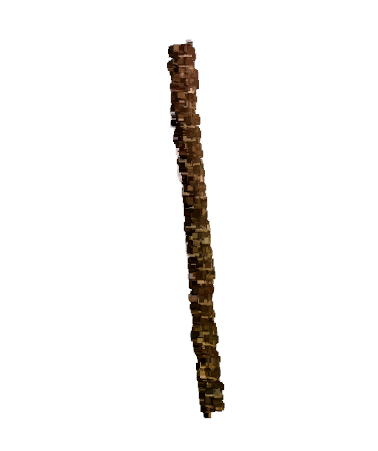}
    \caption{Stalk Point Cloud.}
    \label{fig:point_cloud}
\end{figure}

\subsection*{Stalk Axis Estimation}
Following outlier removal, the central longitudinal axis of the stalk was estimated by applying Principal Component Analysis (PCA) to the filtered 3D point cloud. The analysis used Singular Value Decomposition (SVD) on the centered point coordinates. The first principal component, which represents the direction of maximum variance, was extracted to define the stalk's primary axis. This direction vector was adjusted to point in the positive Z direction to ensure consistent orientation. A 3D line segment was generated along this axis to serve as a visual and computational reference for the subsequent perpendicular slicing of the point cloud (see Fig. 4).

\begin{figure}[htbp]
    \centering
    \includegraphics[width=0.4\textwidth]{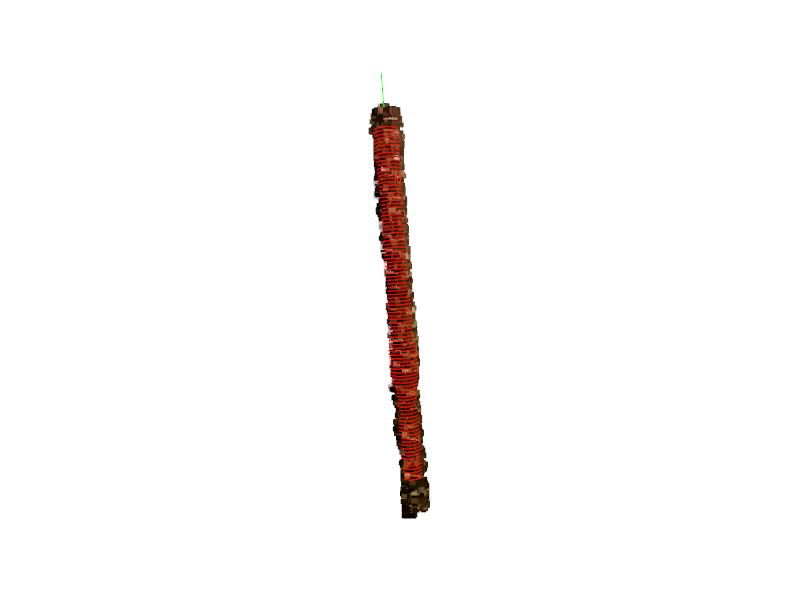}
    \caption{Stalk Point Cloud With Central Axis and Slicing.}
    \label{fig:stalk_axis}
\end{figure}

\subsection*{Volumetric Slicing and Diameter Estimation}
For detailed geometric analysis, the point cloud was partitioned into 100 discrete, parallel slices oriented orthogonally to the estimated longitudinal axis. All points were first projected onto this axis to determine the full extent of the slicing region. To mitigate potential distortions from incomplete data at the extremities, the slicing process was confined to the central 80\% of this projected length, excluding the terminal 10\% at each end. Each slice was defined as a volumetric section with a thickness of \SI{1.5}{\centi\meter}, containing all points within a specified orthogonal distance from a central plane. This method ensured that each cross-section contained a sufficient point density for robust analysis (see Fig. 4).

For each slice:
\begin{itemize}
    \item Points within a fixed distance (\SI{15}{\milli\meter}) of the slicing plane were selected and projected onto a 2D plane orthogonal to the stalk axis.
    \item The projected points were filtered using the DBSCAN clustering algorithm (\( \varepsilon = \SI{0.25}{\milli\meter} \), minimum samples = 5) to remove spatial outliers (\cite{buitinck_api_2013}).
    \item The distances from the slice center to each remaining point were calculated, and the 95th percentile of these distances was used to estimate the radius. This provides a close approximation of the stalk surface while reducing the influence of outliers.
\end{itemize}

\subsection*{Visualization and Output}
All geometric outputs—including filtered point clouds, estimated axes, and fitted cross-sectional profiles—were visualized using Open3D. These visualizations were automatically saved for each processed image pair to enable qualitative inspection and validation of the pipeline's geometric accuracy. Additionally, histograms of retained slice diameters were generated and annotated with corresponding ground-truth values to highlight agreement and variance. All numerical results, including slice-level diameters and summary statistics, were exported to CSV format to support reproducibility, external validation, and further statistical analysis.

\section{Experiments}

\subsection*{Evaluation Metrics}

Four standard statistical metrics were used to assess the performance of the diameter estimation pipeline. These metrics provide a comprehensive view of the model's accuracy, error magnitude, and correlation with ground-truth data. In the following equations, \(y_i\) represents the actual ground-truth value, \(\hat{y}_i\) is the model's predicted value, \(\bar{y}\) is the mean of the actual values, and \(n\) is the total number of samples.

\begin{itemize}
    \item \textbf{Mean Absolute Error (MAE)} was calculated as the average of the absolute differences between predicted diameters and ground-truth measurements. This metric provides a direct, interpretable measure of the average error magnitude in the original units (meters).
    \[
    \text{MAE} = \frac{1}{n} \sum_{i=1}^{n} |y_i - \hat{y}_i|
    \]

    \item \textbf{Mean Absolute Percentage Error (MAPE)} was calculated to assess the average error relative to the magnitude of the actual measurements. As a scale-independent metric, MAPE expresses the mean error as a percentage, providing an intuitive understanding of the model's relative accuracy.
    \[
    \text{MAPE} = \frac{100\%}{n} \sum_{i=1}^{n} \left| \frac{y_i - \hat{y}_i}{y_i} \right|
    \]

    \item \textbf{Root Mean Square Error (RMSE)} was also used to measure model accuracy. By squaring residuals before they are averaged, RMSE disproportionately penalizes larger errors, making it a useful metric for evaluating performance when significant deviations are undesirable.
    \[
    \text{RMSE} = \sqrt{\frac{1}{n} \sum_{i=1}^{n} (y_i - \hat{y}_i)^2}
    \]

    \item \textbf{The coefficient of determination (\(R^2\))} was calculated to quantify the proportion of variance in the actual stalk diameters that is explained by the model's predictions. An \(R^2\) value closer to 1.0 indicates a stronger linear correlation and a better model fit.
    \[
    R^2 = 1 - \frac{\sum_{i=1}^{n} (y_i - \hat{y}_i)^2}{\sum_{i=1}^{n} (y_i - \bar{y})^2}
    \]
\end{itemize}

\subsection*{Experiment Setup}
A set of ten artificial plant stalks with varying diameters was used to validate the pipeline. Artificial stalks were chosen to provide simplified geometries for initial validation and because real crop stalks were seasonally unavailable during development. This approach allowed for a focused assessment of the pipeline's measurement capabilities by minimizing external variables such as dynamic lighting, occlusion, and wind motion.

Data was acquired using a Luxonis OAK-D Short Range (SR) camera. This model was chosen for its suitability in environments with limited space, such as narrow crop rows where a short working distance is necessary. Its \SI{20}{\milli\meter} stereo baseline provides an ideal depth perception range starting at \SI{30}{\centi\meter}, making it effective for accurately measuring nearby objects. For each of the ten plants, an RGB-D image pair was captured from the minimum ideal distance of \SI{30}{\centi\meter}.

Ground-truth diameters were established by averaging three measurements taken along the central portion of each stalk with a digital caliper (\SI{0.01}{\milli\meter} precision). The pipeline's predicted diameters were then compared against these ground-truth values to evaluate performance.

With the algorithm's performance established under controlled conditions, future work will focus on adapting the system for in-field implementation to test its robustness in a real-world environment.

\subsection*{Results}
Table 1 summarizes the calculated slice diameters, predicted stalk diameters, and actual diameters for each of the 10 test plants. Figures 5 through 14 provide visualizations of the 3D analysis for each plant used in the experiment.

\begin{table}[htbp]
    \centering
    \caption{Comparison of predicted vs. actual stalk diameters.}
    \label{tab:results_comparison}
    \small 
    \sisetup{round-mode=places, round-precision=5}
    \begin{tabular}{l S S S S S[table-format=1.5] S[table-format=1.5]}
        \hline
        \makecell{\textbf{Sample}\\\textbf{ID}} & {\makecell{\textbf{Slice 1}\\\textbf{(m)}}} & {\makecell{\textbf{Slice 25}\\\textbf{(m)}}} & {\makecell{\textbf{Slice 50}\\\textbf{(m)}}} & {\makecell{\textbf{Slice 100}\\\textbf{(m)}}} & {\makecell{\textbf{Predicted}\\\textbf{(m)}}} & {\makecell{\textbf{Actual}\\\textbf{(m)}}} \\
        \hline
        rgb0.png & 0.01504 & 0.01497 & 0.01432 & 0.02669 & 0.01517 & 0.01504 \\
        rgb1.png & 0.01581 & 0.01008 & 0.01116 & 0.01540 & 0.01161 & 0.01286 \\
        rgb2.png & 0.01232 & 0.01099 & 0.01217 & 0.01095 & 0.01134 & 0.01220 \\
        rgb3.png & {---}   & 0.01819 & 0.01668 & {---}   & 0.01561 & 0.01568 \\
        rgb4.png & 0.01182 & 0.01141 & 0.01200 & 0.01033 & 0.01155 & 0.01267 \\
        rgb5.png & 0.01610 & 0.01364 & 0.01405 & 0.02484 & 0.01516 & 0.01550 \\
        rgb6.png & 0.05144 & 0.01227 & 0.01372 & 0.01271 & 0.01311 & 0.01304 \\
        rgb7.png & 0.01492 & 0.01538 & 0.01338 & 0.02058 & 0.01530 & 0.01451 \\
        rgb8.png & 0.01573 & 0.01289 & 0.01687 & 0.01473 & 0.01498 & 0.01469 \\
        rgb9.png & 0.01386 & 0.01440 & 0.01636 & 0.01323 & 0.01323 & 0.01275 \\
        \hline
    \end{tabular}
\end{table}

\begin{figure}[htbp]
    \centering
    \begin{subfigure}[b]{0.48\textwidth}
        \centering
        \includegraphics[height=4cm]{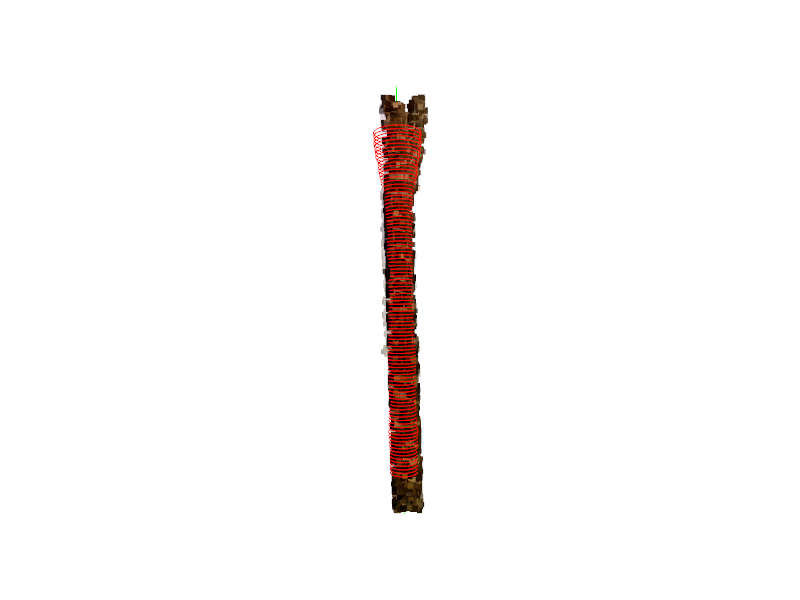}
        \caption{3D Analysis of Stalk 0.}
        \label{fig:3d_analysis_0}
    \end{subfigure}
    \hfill
    \begin{subfigure}[b]{0.48\textwidth}
        \centering
        \includegraphics[height=4cm]{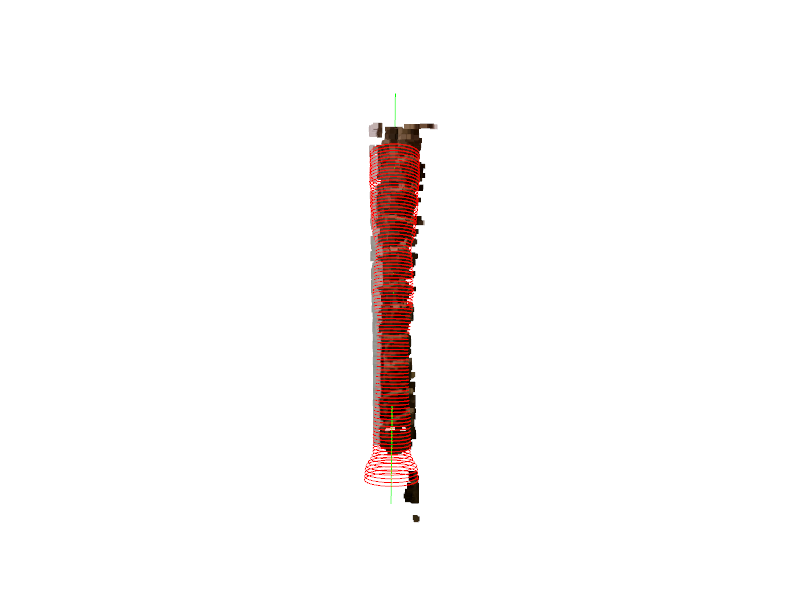}
        \caption{3D Analysis of Stalk 1.}
        \label{fig:3d_analysis_1}
    \end{subfigure}

    \vspace{1em} 

    \begin{subfigure}[b]{0.48\textwidth}
        \centering
        \includegraphics[height=4cm]{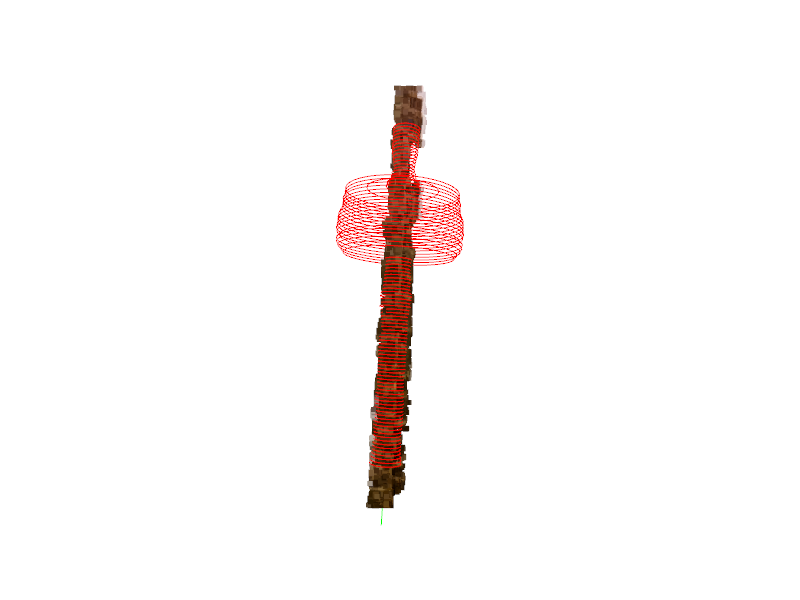}
        \caption{3D Analysis of Stalk 2.}
        \label{fig:3d_analysis_2}
    \end{subfigure}
    \hfill
    \begin{subfigure}[b]{0.48\textwidth}
        \centering
        \includegraphics[height=4cm]{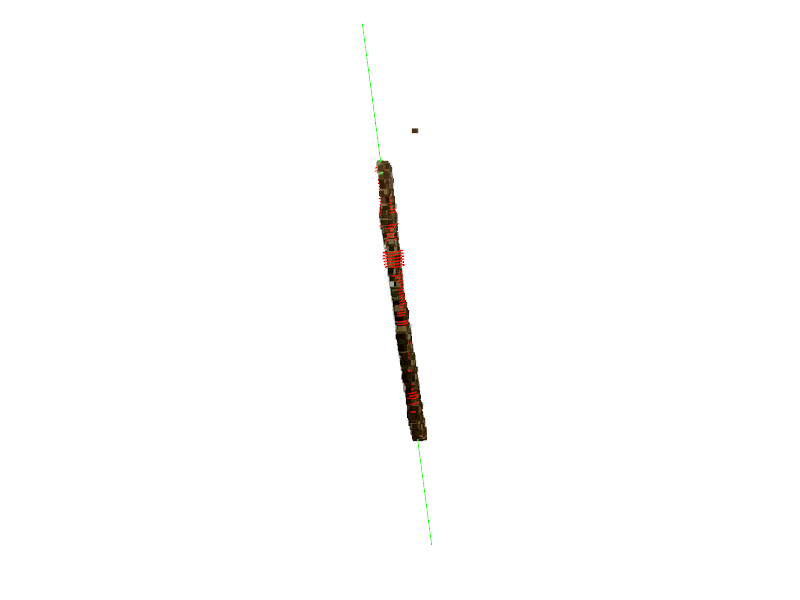}
        \caption{3D Analysis of Stalk 3.}
        \label{fig:3d_analysis_3}
    \end{subfigure}
    
    \caption{Visualizations of the reconstructed 3D point cloud, estimated central axis (red), and volumetric slices for samples 0-3.}
    \label{fig:3d_analysis_group1}
\end{figure}
\begin{figure}[htbp]
    \centering
    \begin{subfigure}[b]{0.48\textwidth}
        \centering
        \includegraphics[height=4cm]{rgb4_viz.png}
        \caption{3D Analysis of Stalk 4.}
        \label{fig:3d_analysis_4}
    \end{subfigure}
    \hfill
    \begin{subfigure}[b]{0.48\textwidth}
        \centering
        \includegraphics[height=4cm]{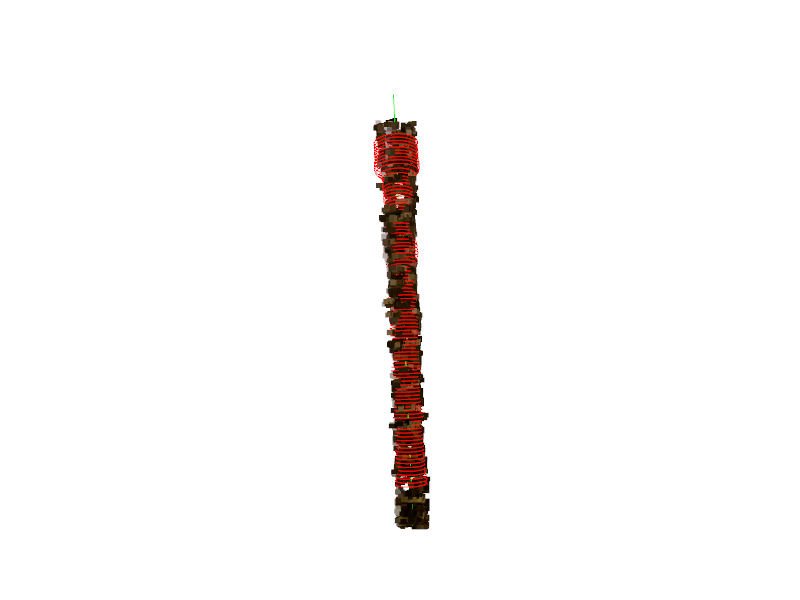}
        \caption{3D Analysis of Stalk 5.}
        \label{fig:3d_analysis_5}
    \end{subfigure}

    \vspace{1em} 

    \begin{subfigure}[b]{0.48\textwidth}
        \centering
        \includegraphics[height=4cm]{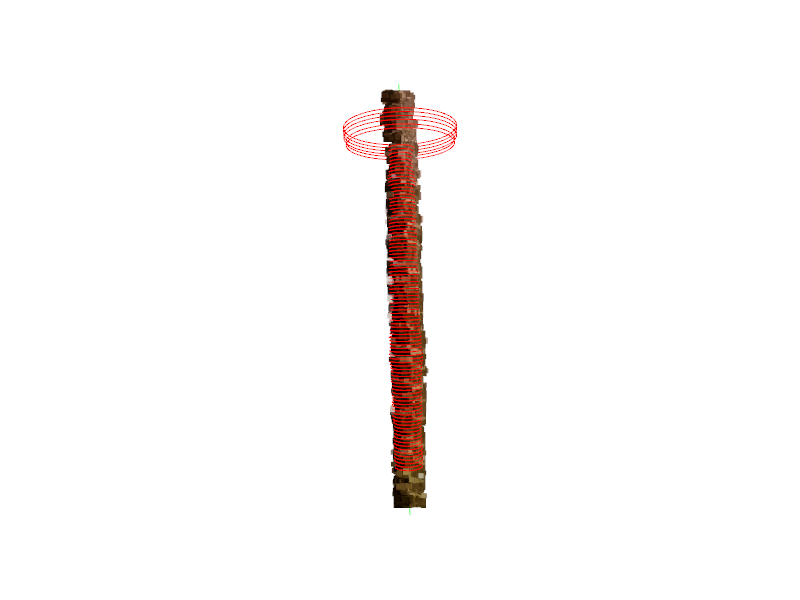}
        \caption{3D Analysis of Stalk 6.}
        \label{fig:3d_analysis_6}
    \end{subfigure}
    \hfill
    \begin{subfigure}[b]{0.48\textwidth}
        \centering
        \includegraphics[height=4cm]{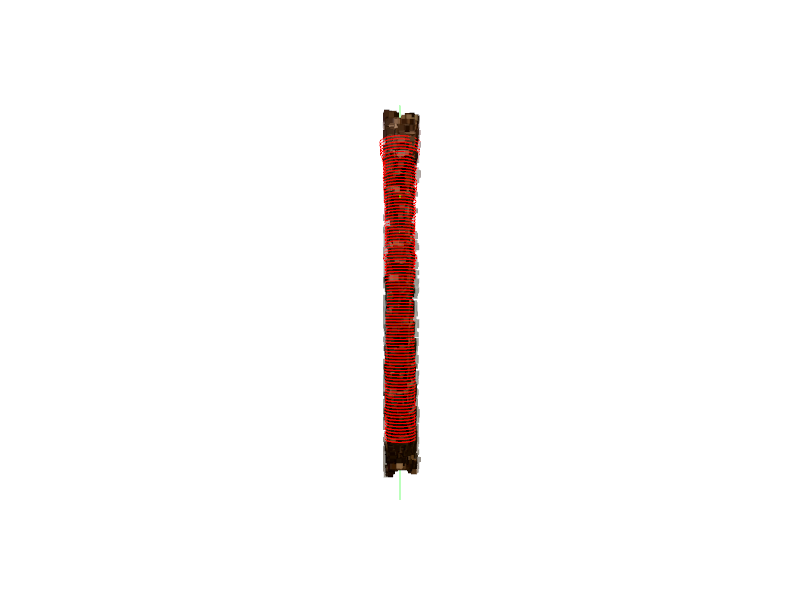}
        \caption{3D Analysis of Stalk 7.}
        \label{fig:3d_analysis_7}
    \end{subfigure}
    
    \caption{Visualizations of the reconstructed 3D point cloud, estimated central axis (red), and volumetric slices for samples 4-7.}
    \label{fig:3d_analysis_group2}
\end{figure}
\begin{figure}[htbp]
    \centering
    \begin{subfigure}[b]{0.48\textwidth}
        \centering
        \includegraphics[height=4cm]{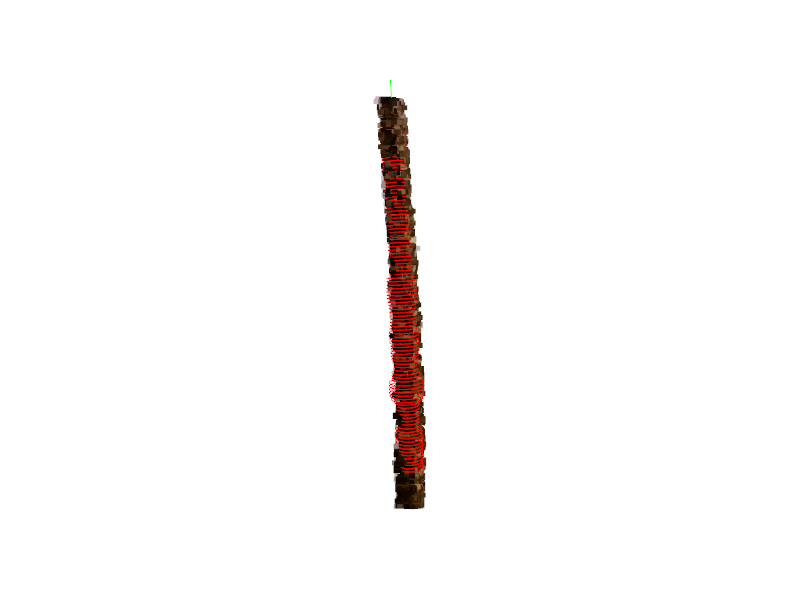}
        \caption{3D Analysis of Stalk 8.}
        \label{fig:3d_analysis_8}
    \end{subfigure}
    \hfill
    \begin{subfigure}[b]{0.48\textwidth}
        \centering
        \includegraphics[height=4cm]{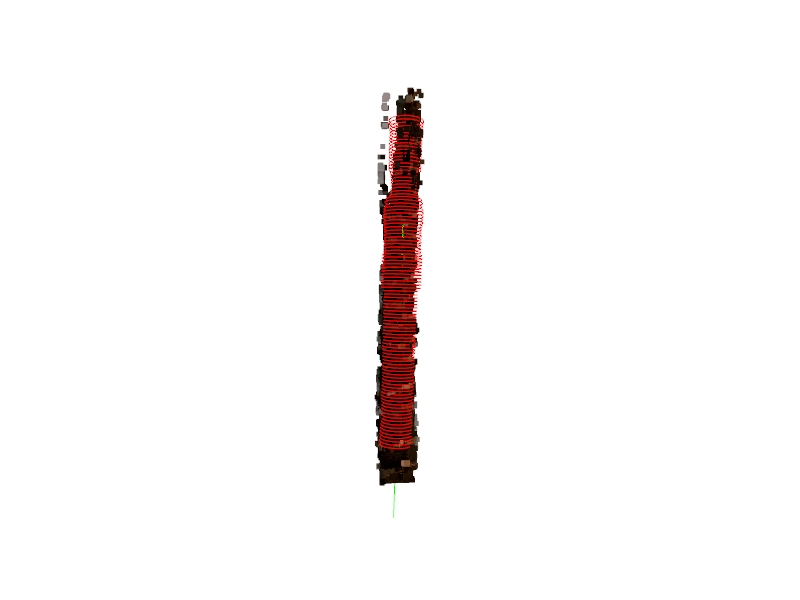}
        \caption{3D Analysis of Stalk 9.}
        \label{fig:3d_analysis_9}
    \end{subfigure}
    
    \caption{Visualizations of the reconstructed 3D point cloud, estimated central axis (red), and volumetric slices for samples 8-9.}
    \label{fig:3d_analysis_group3}
\end{figure}

\section{Discussion}

\subsection*{Accuracy}

The pipeline demonstrated promising performance in estimating stalk diameters from RGB-D data, achieving sub-millimeter precision on artificial plant stalks in a controlled setting. The mean absolute error (MAE) of \SI{0.539}{\milli\meter} and root mean square error (RMSE) of \SI{0.681}{\milli\meter} (Table~\ref{tab:performance_metrics}) suggest a high degree of agreement between predicted and ground-truth diameters, especially considering the small scale of the target structures. The mean absolute percentage error (MAPE) of 4.08\% further supports the method's suitability for tasks requiring fine-grained morphological measurement.

\begin{table}[htbp]
    \centering
    \caption{Overall model performance metrics based on the validation set.}
    \label{tab:performance_metrics}
    \begin{tabular}{lr}
        \hline
        \textbf{Metric} & \textbf{Value} \\
        \hline
        Mean Absolute Error (MAE) & \SI{0.539}{\milli\meter} \\
        Mean Absolute Percentage Error (MAPE) & 4.08\% \\
        Root Mean Square Error (RMSE) & \SI{0.681}{\milli\meter} \\
        Coefficient of Determination (\(R^2\)) & 0.7020 \\
        \hline
    \end{tabular}
\end{table}

The coefficient of determination ($R^2 = 0.7020$) indicates that the model's predictions explain approximately 70\% of the variance in the actual diameter values. While this value suggests a good fit, it should be interpreted in the context of the test dataset; the use of artificial stalks resulted in limited diameter variability across samples, which can artificially deflate the $R^2$ metric. In datasets with broader structural diversity, the same level of absolute accuracy would likely yield a higher $R^2$ score.

Several samples (e.g., \texttt{}, \texttt{rgb3.png}, \texttt{rgb8.png}) exhibited close agreement between predicted and actual values, with absolute errors under \SI{0.2}{\milli\meter} (Table~\ref{tab:results_comparison}). This demonstrates the pipeline's ability to produce high-fidelity measurements when slice quality is consistent.

Variability in individual slice measurements—especially at the extremities such as Slice 100—reflects localized fluctuations likely caused by 3D noise or segmentation drift near stalk edges. For instance, \texttt{rgb0.png} showed a sharp diameter increase at Slice 100 (\SI{26.69}{\milli\meter}), which appears to be an outlier relative to the stalk's overall profile. These discrepancies underscore the importance of robust filtering, which was addressed in the pipeline through PCA-based axis estimation and statistical refinement of the slice data.

\subsection*{Ablation Study}

A comprehensive ablation study was conducted to validate the contribution of each major component to the proposed pipeline. The complete pipeline, which serves as our baseline, incorporates Statistical Outlier Removal (SOR), DBSCAN slice filtering, a 95th percentile circle fitting method, and aggregation of slice diameters within one standard deviation (1-std) of the mean. The performance of each configuration is summarized in Table~\ref{tab:ablation_results}.

\begin{table}[htbp]
    \centering
    \caption{Ablation study results comparing pipeline component performance}
    \label{tab:ablation_results}
    \small 
    \begin{tabular}{
        l 
        S[table-format=1.6]
        S[table-format=2.2]
        S[table-format=1.6]
        S[table-format=-2.4]
    }
        \hline
        {\makecell[l]{\textbf{Configuration}}} & 
        {\makecell{\textbf{MAE}\\\textbf{(m)}}} & 
        {\makecell{\textbf{MAPE}\\\textbf{(\%)}}} & 
        {\makecell{\textbf{RMSE}\\\textbf{(m)}}} & 
        {\makecell{\textbf{R²}}} \\
        \hline
        \textbf{Baseline (Full Pipeline)} & \bfseries 0.000539 & \bfseries 4.08 & \bfseries 0.000681 & \bfseries 0.7020 \\
        No 1-std Aggregation (Mean) & 0.001596 & 12.19 & 0.002377 & -2.6302 \\
        No DBSCAN Slice Filtering & 0.001542 & 10.57 & 0.002035 & -1.6600 \\
        No Statistical Outlier Removal & 0.000537 & 4.07 & 0.000682 & 0.7013 \\
        Circle Fit Method: Mean & 0.006152 & 44.66 & 0.006197 & -23.6603 \\
        Circle Fit Method: Median & 0.006499 & 47.29 & 0.006557 & -26.6169 \\
        \hline
    \end{tabular}
\end{table}


\begin{table}[htbp]
    \centering
    \caption{Component importance ranking. The MAE Impact is the increase in Mean Absolute Error (MAE) when a pipeline component is removed or replaced.}
    \label{tab:component_importance}
    \small
    \begin{tabular}{
        l
        c
        S[table-format=1.6]
    }
        \hline
        {\makecell[l]{\textbf{Component}}} & 
        {\makecell{\textbf{Importance}\\\textbf{Rank}}} & 
        {\makecell{\textbf{MAE Impact}\\\textbf{(m)}}} \\
        \hline
        Circle Fit Method(95th percentile vs. others) & 1 & 0.005960 \\
        1-std Aggregation                             & 2 & 0.001057 \\
        DBSCAN Slice Filtering                        & 3 & 0.001003 \\
        Statistical Outlier Removal(SOR)             & 4 & -0.000002 \\
        \hline
    \end{tabular}
\end{table}

From these results, it is clear that the choice of \textbf{circle fitting, result aggregation, and DBSCAN-based filtering} methods are critical components for achieving high precision.

\textbf{Circle Fitting Method:} The most significant performance degradation occurred when the circle fitting method was changed from the 95th percentile to either the mean or median radius. Using these alternatives resulted in a drastic reduction in accuracy, with the MAE increasing by more than tenfold and the $R^2$ value becoming strongly negative (-23.66 and -26.61, respectively). This indicates that the raw point cloud slices contain significant noise and outliers that disproportionately affect central tendency metrics. The 95th percentile method proves to be a robust estimator, as it effectively ignores these inner noise points and fits to the outer boundary of the stalk, which is crucial for an accurate diameter measurement.

\textbf{Result Aggregation:} The second most critical component was the final aggregation method. When the \texttt{1-std aggregation} filter was disabled in favor of a simple mean of all slice diameters, the MAE nearly tripled, and the $R^2$ value dropped to -2.63. This is because outlier slices, likely from the ends of the stalk or areas with occlusions, can skew the overall average. By filtering to include only measurements within one standard deviation of the mean, our pipeline effectively removes these erroneous estimates, leading to a much more reliable and accurate final diameter.

\textbf{Slice-level Filtering (DBSCAN):} Removing DBSCAN-based filtering from each 2D slice also caused a substantial drop in performance, with the MAE increasing to \SI{0.001542}{m} and $R^2$ becoming negative. Although not as detrimental as changing the fitting method, this result shows the importance of cleaning data at the slice level before fitting. DBSCAN effectively removes disparate points that are not part of the main stalk cross-section, preventing them from influencing the percentile-based radius calculation.

\textbf{Point Cloud Filtering (SOR):} Interestingly, removing the initial Statistical Outlier Removal (SOR) filter had a negligible effect on the final metrics, with performance nearly identical to the baseline (MAE of 0.000537 vs. 0.000539). This suggests that while SOR cleans the overall point cloud, the subsequent slice-level filtering (DBSCAN) and robust statistical methods (95th percentile fit and 1-std aggregation) are powerful enough to mitigate the noise on their own. For this specific application, the initial, computationally expensive SOR step appears to be redundant.

This ablation study confirms that the robustness of our pipeline is primarily derived from its statistical techniques for circle fitting and result aggregation, which are specifically designed to handle the noisy data produced by depth sensors.

\section{Conclusion}

\textbf{Summary of the System:} This study developed and validated a pipeline for estimating plant stalk diameter from RGB-D images. The system uses a YOLOv11 model to segment individual stalks, determines the true axis of each stalk with Principal Component Analysis (PCA), calculated by Singular Value Decomposition (SVD), and then employs a volumetric slicing method to derive precise diameter measurements.

\textbf{Strengths:} The primary strength of this pipeline is its high accuracy, achieving sub-millimeter precision (MAE = \SI{0.539}{\milli\meter}) on the test dataset. The ablation study demonstrated that the system's robustness stems from its statistical methods—specifically the 95th percentile circle fit and 1-std aggregation—which effectively handle sensor noise. This makes the initial, computationally expensive SOR filter redundant, an important finding for future optimization.

\textbf{Limitations:} This study's validation was conducted in a controlled lab environment using artificial stalks. This approach was a crucial first step, allowing the core measurement methodology to be proven without confounding variables from a field environment, such as dynamic lighting, wind-induced motion, and partial occlusion. While this successfully established the pipeline's geometric measurement capability, its performance in an operational agricultural setting is the next key validation step.

\textbf{Future Work:} Future work will focus on transitioning this system from controlled laboratory settings to real-world field environments. The immediate next step involves testing the pipeline on actual crops such as maize or sorghum, which will necessitate training a more robust segmentation model on a diverse, in-field dataset. To enable real-time deployment, we will also explore alternatives to deep segmentation networks, including polynomial-based detection methods as proposed in \cite{RowDetr}, or lightweight edge-based approaches such as \cite{diffedge}. These efforts aim to establish the proposed system as a practical and scalable solution for high-throughput phenotyping in agricultural settings.

\newpage
\printbibliography
\end{document}